\documentclass[letterpaper]{article}
\usepackage{aaai}
\usepackage{times}
\usepackage{helvet}
\usepackage{courier}
\usepackage{graphicx}
\usepackage{subfigure}
\usepackage{xcolor}
\usepackage{booktabs}
\usepackage{multirow}
\usepackage{sidecap}
\usepackage{amsthm}
\usepackage{amsmath}
\usepackage{amsfonts}
\usepackage{url}
\allowdisplaybreaks 
\def\Pr{\mathop{\rm Pr}\nolimits}

\newcommand{\eat}[1]{}

\newcommand{\namecite}[1]{\citeauthor{#1} \shortcite{#1}}

\theoremstyle{definition}

\newcommand\truefalse{T/F }
\newcommand\kb{\mathit{KB}}

\newcommand\setup{\mathit{setup}}
\newcommand\query{\mathit{query}}
\newcommand\antecedent{\mathit{antecedent}}
\newcommand\consequent{\mathit{consequent}}

\newcommand\isa{\mathit{isa}}
\newcommand\entails{\mathit{entails}}
\newcommand\agent{\mathit{agent}}
\newcommand\object{\mathit{object}}
\newcommand\cause{\mathit{cause}}
\newcommand\effect{\mathit{effect}}
\newcommand\enables{\mathit{enables}}
\newcommand\function{\mathit{function}}

\newcommand\sameas{\mathit{sameAs}}

\newcommand\node{\mathit{node}}
\newcommand\edge{\mathit{edge}}
\newcommand\holds{\mathit{holds}}
\newcommand\aligns{\mathit{aligns}}
\newcommand\inlhs{\mathit{inLhs}}
\newcommand\inrhs{\mathit{inRhs}}
\newcommand\lhsholds{\mathit{lhsHolds}}
\newcommand\rhsholds{\mathit{rhsHolds}}
\newcommand\proves{\mathit{proves}}
\newcommand\ruleProves{\mathit{ruleProves}}

\nocopyright   % no AAAI copyright for StarAI submission

\begin{document}

\title{Markov Logic Networks for Natural Language Question Answering}
%\title{Probabilistic Inference for Natural Language Question Answering}
%\title{Lexically Enhanced Probabilistic Inference\\ for Natural Language Question Answering}
%\title{Probabilistic Inference with Lexical Semantics\\ for Natural Language Question Answering}

%Adapting MLNs for Question Answering
%Efficient 
%Representation and Efficiency (?) Challenges
%Formulating QA in an MLN Framework: Investigating/Addressing/Attacking/Overcoming Rep. and Effi. Chal.

\author{
  Tushar Khot\\
  Allen Institute for AI\\
  Seattle, WA 98103
\And
  Niranjan Balasubramanian\\
  Dept.~of Computer Science\\
  Stony Brook University\\
  Stony Brook, NY 11794
\And
  Eric Gribkoff\\
  Computer Sci.~and Engr.\\
  University of Washington\\
  Seattle, WA 98195
\And
  Ashish Sabharwal\\
  {\bf \large Peter Clark}\\
  {\bf \large Oren Etzioni}\\
  Allen Institute for AI\\
  Seattle, WA 98103
}

\maketitle

%%%%%%%%%%%%%%%%%%%%%
%% ABSTRACT
%%%%%%%%%%%%%%%%%%%%%

\begin{abstract}
Our goal is to answer elementary-level science questions using knowledge extracted
automatically from science textbooks, expressed in a subset of first-order
logic. Given the incomplete and noisy nature of these automatically extracted
rules, Markov Logic Networks (MLNs) seem a natural model to use, but the exact
way of leveraging MLNs is by no means obvious. We investigate three ways of
applying MLNs to our task. In the first, we simply use the
extracted science rules directly as MLN clauses. Unlike typical MLN applications, our
domain has long and complex rules, leading to an unmanageable number of
groundings. We exploit the structure present in hard constraints to improve
tractability, but the formulation remains ineffective. In the
second approach, we instead interpret science rules as describing prototypical
entities, thus mapping rules directly
to grounded MLN assertions, whose constants are then clustered using
existing entity resolution methods. This drastically simplifies the network,
but still suffers from brittleness. Finally, our
third approach, called Praline, uses MLNs to align the lexical elements as well as define and control how inference should be performed in this task. Our
experiments, demonstrating a 15\% accuracy boost and a 10x reduction in runtime, suggest that the flexibility
and different inference semantics of Praline are a better fit for the natural
language question answering task.
%while retaining a formal specification of those semantics in the MLN.
%Our experiments suggest that this third approach
%is the most effective for handling natural language inference in our context.
\end{abstract}

%%%%%%%%%%%%%%%%%%%%%
%% INTRODUCTION
%%%%%%%%%%%%%%%%%%%%%
\vspace{-1em}
\section{Introduction}

Many question answering or QA tasks require the ability to reason with knowledge extracted from text. 
We consider the problem of answering questions in standardized science exams~\cite{chb2013:akbc}. In particular, we focus on a subset that tests students' understanding of various kinds of general rules and principles (e.g., \emph{gravity pulls objects towards the Earth}) and their ability to apply these rules to reason about specific situations or scenarios (e.g., \emph{which
force is responsible for a ball to drop?}).

This task can be viewed as a natural first-order reasoning problem specified over general truths expressed over classes of events or entities. However, this knowledge is automatically derived from appropriate science texts. 

In order to effectively reason over knowledge derived from text, a QA system must handle incomplete and potentially noisy knowledge, and reason under uncertainty. Markov Logic Network (MLN) is a formal probabilistic inference framework that allows for robust inference using rules expressed in probabilistic first-order logic~\cite{richardson2006markov}. MLNs have been widely adopted for many tasks~\cite{singla2006er,kok2008er,usp}. Recently, Beltagy et al.~(\citeyear{beltagy2013montague,beltagy2014efficient}) have shown that MLNs can be used to reason with rules derived from natural language. 

While MLNs appear a natural fit, it is \textit{a priori} unclear how to effectively formulate the QA task. Moreover, the unique characteristics of this domain pose new challenges in grounding and ability to control inference under incomplete evidence. 
We investigate two standard formulations, uncover efficiency and brittleness issues, and propose an enhanced formulation more suitable for this domain.
%We investigate the use of MLNs for a QA task that works with noisy knowledge extracted automatically from text. While MLNs appear a natural fit, unique characteristics of the QA domain pose new challenges. We propose three formulations of QA as a marginal inference problem. 
This enhanced formulation, called Praline, significantly outperforms our other MLN formulations, reducing runtime by 10x and improving accuracy by 15\%.

%%%%%%%%%%%%%%%%%%%%%
%% BACKGROUND (AKBC, ETC.)
%%%%%%%%%%%%%%%%%%%%%
\vspace{-0.5em}
\section{Setup: Question Answering Task}

Following \namecite{clark2014:akbc}, we formulate QA as a reasoning task over knowledge derived from textual sources. Specifically, a multiple choice question with $k$ answer options is turned into $k$ true-false questions, each of which asserts some known facts, referred to as the $\setup$, and posits a $\query$. The reasoning task is to determine whether the $\query$ is true given the $\setup$ and the input knowledge. 

The input knowledge is derived from 4th-grade level science texts and augmented with a web search for terms appearing in the texts. Much of this knowledge is in terms of generalities, expressed naturally as IF-THEN rules. We use the representation and extraction procedures of \namecite{clark2014:akbc}, recapitulated briefly here for completeness.

\textbf{Rule Representation:}
The generalities in text convey information about classes of entities and events. Following the neo-davidsonian 
reified representation \cite{boxer2007}, we encode information about events (e.g, falling, dropping etc.) and entities (e.g., ball, stone etc.) using variables. Predicates such as $\agent, \cause, \function, \mathit{towards}, \mathit{in}$ etc., define semantic relationships between the variables. Rather than committing to a type ontology, the variables are associated with their original string representation using an $\isa$ predicate.
%This fixed vocabulary of context-independent predicates, along with general physical relations such as $\mathit{in}, \mathit{towards},$ and $\mathit{with},$ suffice to encode typical relationships among objects in elementary-level science.

The ``if" or $\antecedent$ part of the rule is semantically interpreted as being universally quantified (omitted below for conciseness) whereas every entity or event mentioned only in the ``then" or $\consequent$ part of the rule is treated as existentially quantified. Both $\antecedent$ and $\consequent$ are interpreted as conjunctions.  E.g., ``Growing thicker fur in winter helps some animals to stay warm" translates into:
\begin{align}
  & \isa(g,\text{grow}), \isa(a,\text{some\_animals}), \isa(f,\text{thicker\_fur}), \nonumber\\
  & \isa(w,\text{the\_winter}), \agent(g,a), \object(g,f), \mathit{in}(g,w)\nonumber\\
  & \Rightarrow \exists s, r: \isa(s,\text{stays}), \isa(r,\text{warm}),\nonumber\\
  & \ \ \ \ \ \ \ \enables(g,s), \agent(s,a), \object(s,r)
  \label{rule:fox}
\end{align}

\textbf{Question Representation:}
The question representation is computed similarly except that we use fixed constants (represented as block letters) rather than variables.  E.g., consider the question: ``A fox grows thick fur as the season changes. This helps the fox to (a) hide from danger (b) attract a mate
(c) find food (d) keep warm?''  The \truefalse question corresponding to option (d) translates into:
\begin{align*}
\setup : & \isa(F,\text{fox}), \isa(G,\text{grows}), \isa(T,\text{thick\_fur}),\nonumber\\
         & \agent(G,F), \object(G,T)\nonumber\\
\query : & \isa(K,\text{keep\_warm}), \enables(G,K), \agent(K,F)
\end{align*}

\textbf{Lexical Reasoning:}
Since entities and event variables hold textual values, reasoning must accommodate the lexical variability and textual entailment. For example, the surface forms ``thick$\_$fur" in the question and ``thicker$\_$fur" are semantically equivalent. Also, the string ``fox" entails ``some\_animal". We use a lexical reasoning component based on textual entailment to establish lexical equivalence or entailment between variables. 

\textbf{Most Likely Answer as Inference:}
Given the input KB rules and the question,
we formulate a probabilistic reasoning problem by adding lexical reasoning probabilities and incorporating uncertainties in derived rules. Specifically,
given setup facts $S$ and $k$ answer options $Q_i$, we seek the most likely answer option: $\arg \max_{i \in \{1,\ldots,k\}} \Pr[Q_i \mid S, \kb]$. This is a Partial MAP computation, in general \#P-hard \cite{park02:map-approx}. Hence methods such as Integer Linear Programming are not directly applicable.

%%%%%%%%%%%%%%%%
\subsection{Challenges}

Reasoning with text-derived knowledge presents, in addition to lexical uncertainty, challenges that expose the {\em brittleness} and {\em rigidity} inherent in pure logic-based frameworks. In particular, text-derived rules are incomplete and include lexical items as logical elements, making rule application in a pure logical setting extremely brittle: Many relevant rules cannot be applied because their pre-conditions are not fully satisfied due to poor alignment. For example, naive application of rule (\ref{rule:fox}) on $\setup$ would not conclude $\query$ as the rule requires ``in the winter'' to be true. A robust inference mechanism must allow for rule application with partial evidence.

%To effectively reason with partial matches inference mechanisms must allow for greater flexibility in deciding which partial matches can lead to more confident inferences.
Further, a single text-derived rule may be insufficient to answer a question. E.g., ``Animals grow thick fur in winter" and ``Thick fur helps keep warm" may need to be chained.
%The inference mechanism for QA thus needs to leverage multiple KB rules.

%%%%%%%%%%%%
\vspace{-0.5em}
\section{Probabilistic Formulations}

Statistical Relational Learning (SRL) models~\cite{srlBook} are a natural fit for QA. They provide probabilistic reasoning over knowledge represented in first-order logic, thereby handling uncertainty in lexical reasoning and incomplete matching. While there are many SRL formalisms including Stochastic Logic Programs (SLPs) \cite{slp}, ProbLog \cite{problog}, PRISM \cite{prism}, etc., we use 
Markov Logic Networks (MLNs) for their ease of specification and their ability to naturally handle potentially cyclic rules. We explore three formulations:
 
 {\bf a) First-order MLN:} Given a question and relevant first-order KB rules, we convert them into an MLN program and let MLN inference automatically handle rule chaining. While a natural first-order formulation of the QA task, this struggles with long conjunctions and existentials in rules, as well as relatively few atoms and little to no symmetries. This results in massive grounding sizes, not remedied easily by existing solutions such as
%\emph{lazy} ~\cite{singla2006memory}, \emph{lifted}~\citeeg{gogate2011:ptp,venugopal2012:lbg}, or \emph{structured inference}~\citeeg{domingos2012tractable}. 
lazy, lifted, or structured inference.
We exploit the structure imposed by hard constraints to vastly simplify groundings and bring them to the realm of feasibility, but performance remains poor.

{\bf b) Entity Resolution MLN:} 
%To address efficiency issues, we discard class-based first-order variables in favor of rules with prototypical constants to represent generalities. 
Instead of reasoning with rules that express generalities over classes of individuals, we replace the variables in the previous formulation with prototypical constants. This reduces the number of groundings, while retaining the crux of the reasoning problem defined over generalities. Combining this idea with existing entity resolution approaches %\cite{singla2006er,kok2008er}
substantially improves scalability. However, this turns out to be too brittle in handling lexical mismatches, e.g., different sentence parse structures.

{\bf c) Praline MLN:} Both of the above MLNs rely on exactly matching the relations in the KB and question representation, making them too brittle for this task. In response, PRobabilistic ALignment and INferencE (Praline) performs inference using primarily the string constants but guided by the edge structure. We relax the rigidity in rule application by explicitly modeling the desired QA inference behavior.  %and using fine-grained control over the degree of alignment between lexical elements and its influence on inference. \tushar{combine!}

%
%The Praline MLN formulation instead performs inference using only the string constants but guided by the relational edge structure of the underlying knowledge graph.  We leverage first-order logic to define the inference behavior using a static MLN common to all questions, which can now be easily modified to control alignment and inference behavior.
% Eric: thought it read better this way, feel free to restore
%To effectively address these challenges we investigate three different Markov Logic Networks (MLNs) formulations with varying capabilities [TODO]. While MLNs have been applied successfully to NLP tasks, it is not clear a priori if they would be suitable for reasoning with automatically derived knowledge.
%%%%%%%%%%%%%%%%%%%%%
%% FIRST-ORDER MLN
%%%%%%%%%%%%%%%%%%%%%
%\vspace{-0.5em}
\subsection{(A) First-Order MLN Formulation}
%While converting FOL rules capturing KB and lexical knowledge into an
%MLN program may seem straightforward at first, ensuring that the
%generated program is semantically meaningful and supports scalable
%inference requires great care. We next discuss details of the MLN
%generation component of our architecture, describe various design
%decisions, and explain how we efficiently transform the natural but
%complex syntactic structure of our rules into a simpler form that can
%be fed to existing MLN solvers.
%
%An MLN program can be thought of as a first-order template for
%building ground Markov Networks. It is defined by the following
%components, which we discuss in the context of our QA setting.

For a set $R$ of first-order KB rules, arguably the most natural way to represent the QA task of computing $\Pr[Q_i \mid S,R]$ as an MLN program $M$ is to simply add $R$ essentially verbatim as first-order rules in $M$. Quantified variables of $M$ are  those occurring in $R$. Constants of $M$ are the string representations (e.g., ``fox", ``thicker\_fur") in $Q_i, S,$ and $R$, as well as the constants in the $Q_i$ and $S$ (e.g., $F$, $T$). In addition, for all existentially quantified variables, we introduce a new domain constant. Predicates of $M$ are those in $R$, along with a binary $\entails$ predicate representing the lexical entailment blackbox, which allows $M$to probabilistically connect lexically related constants such as ``thick\_fur" and ``thicker\_fur" or ``fox" and ``some\_animals". $\entails$ is defined to be closed-world and is not necessarily transitive.

%\paragraph{Types:}
\textbf{Refined Types}:
For improved semantics and reduced grounding size, $M$ has entities (A), events (E), and strings as three basic types, and predicates of $M$ are appropriately typed (e.g., $\mathit{agentEA}$, $\mathit{entailsEE}$, $\mathit{entailsAA}$). Further, we avoid irrelevant groundings by using \emph{refined types} determined dynamically: if $r(x,y)$ appears only with constants associated with strings
%$T = \{\text{person}, \text{man}\}$ 
$T$ as the second argument, then $M$ contains $r(x,y) \to !\isa(y,s)$ for all strings $s$ with a zero entailment score with all strings in $T$. 
%This prevents irrelevant groundings, such as $\agent(x,\text{Ball})$, from being created when $\text{Ball}$ is a constant associated with the string ``ball".
%\eric{Question: the refined types add a rule saying agent(x,y) => !isa(y,s); when I read the text it's not clear how the y in agent and the y in isa are connected.} 
%\ashish{Repharsed and corrected.}

% The (finite) domains of entity and event variables consist of the corresponding non-string constants that appear in the $\setup$ of the question. In addition, for each existentially quantified variable in the MLN rules, we introduce a new domain constant of the appropriate type. The domain for the String type consists of all strings appearing in the question and the KB rules.

%\paragraph{Input predicates:} 
\textbf{Evidence}:
Soft evidence for $M$ consists of $\entails$ relations between every ordered pair of entity (or event) strings, e.g., $\entails(\text{fox}, \text{some\_animals})$. Hard evidence for $M$ comprises grounded atoms in $S$. %In addition, if $Q_i$ contains $\isa(A,B)$ as well as another relation with $A$, we include $\isa(A,B)$ as hard evidence as it is unnecessary to consider a world where $A$ is not tied to the string $B$.

\textbf{Query}:
%The relation predicates (e.g. $\mathit{agentEA}$) are the open-world predicates in the MLNs.
The query atom in $M$ is $\mathit{result}()$, a new zero-arity predicate $\mathit{result}()$ that is made equivalent to the conjunction of the predicates in $Q_i$ that have not been included in the evidence. We are interested in computing $\Pr[\mathit{result}()=\text{true}]$.

%\paragraph{Rules:} 
\textbf{Semantic Rules}: 
In addition to KB science rules, we add
\emph{semantic rules} that capture the intended meaning of our predicates,
such as every event has a unique agent, $\cause(x,y) \to \effect(y,x)$, etc. Semantic predicates also enforce natural restrictions
such as non-reflexivity, $!r(x,x)$, and
anti-symmetry, $r(x,y) \to !r(y,x)$.
%A world violating any such semantic constraint is deemed to be in feasible.
%We also exploit the closed-world assumption on the $\entails$ predicate to make the $\isa$ relation effectively closed world: $\forall x,y,z: \isa(x,y), \isa(x,z) \to \entails(y,z)$.

%\footnote{Our implementation also allows one-step-removed entailments: in order to account for situations such as ``red ball'' is ``red'' and red ball'' is ``ball'', but $\entails(\text{red},\text{ball})$ is false.}

% As we will discuss later, the presence of this constraint as well as other
% hard semantic constraints play a crucial role in being able to
% successfully apply MLNs to the QA task.

Finally, to help bridge lexical gaps more, we use a simple external lexical alignment algorithm to estimate how much does the $\setup$ entail $\antecedent_r$ of each KB rule $r$, and how much does $\consequent_r$ entail $\query$. These are then added as two additional MLN rules per KB rule.

These rules have the following first-order logic form:
\begin{small}
\begin{align*}
  \forall x_1,..,x_k \bigwedge_i R_i(x_{i_1},x_{i_2}) \rightarrow \exists x_{k+1},..,x_{k+m} \bigwedge_j R_j(x_{j_1},x_{j_2})
\end{align*}
\end{small}
%
%Such a rule can be converted to a number of ground constraints by
%grounding each of its constituent predicates to atoms.
%The lexical alignment rules $S \to L_j$ and $R_j \to Q_i$ also have
%the same general FOL form, where existentially quantified variables in
%the consequent are precisely those that do not appear in the
%corresponding antecedent.
%
\emph{Existentials spanning conjunctions} in the
consequent of this rule form can neither be directly fed into existing MLN systems nor
efficiently expanded naively into a standard conjunctive normal form
(CNF) without incurring an exponential blowup during the
transformation. To address this, we introduce a new
``existential'' predicate $E_j^\alpha(x_1, \ldots, x_k, x_{k+j})$ for
each existential variable $x_{k+j}$ in each such rule $\alpha$. This
predicate becomes the consequent of $\alpha$, and subsequent hard
MLN rules make it equivant to the original consequent.

\subsubsection{Boosting Inference Efficiency.}

A bottleneck in using MLN solvers out-of-the-box for this QA formulation is the prohibitively large grounded network size. For instance, 31\% of our runs that timed out during the MLN grounding phase after 6 minutes were dealing, on average, with $1.4 \times 10^6$ ground clauses. Such behavior has also been observed, perhaps to a lesser degree, in related NLP tasks~\cite{beltagy2014efficient,beltagy2014sts}.
%such as RTE \cite{beltagy2014efficient} and STS \cite{beltagy2014sts}.

%\ashish{SHRINK THIS PARAGRAPH...}
%\eric{shrunken, but still pretty big}

Large grounding size is, of course, a well-studied problem in the MLN
literature. However, a key difference from previously studied MLNs
is that our QA encodings have small domain sizes and, therefore,
\emph{very few ground atoms} to start with. Existing techniques for
addressing the grounding challenge were inspired by
the unmanageable number of ground atoms often seen in traditional MLN
applications, and work by grouping them into large classes of interchangeable atoms~\cite{salvobraz2005:lifted,gogate2011:ptp,venugopal2012:lbg,domingos2012tractable,niepert2014tractability}.
%(e.g., a domain size of $1000$ resulting in $10^6$ groundings of the $\mathit{Friend}(x,y)$ relation). For instance, both Lifted Inference \cite{salvobraz2005:lifted,gogate2011:ptp,niepert2014tractability} and Tractable Markov Logic (TML) \cite{domingos2012tractable} exploit the fact that ground atoms can sometimes be grouped into large classes of mutually interchangeable atoms. 
Similarly, memory-efficient Lazy Inference
\cite{singla2006memory} and FROG~\cite{shavlik2009speeding}
focus only on relevant atoms.

These methods were ineffective on our MLNs. E.g., lazy inference in Alchemy-1 reduced $\sim$70K ground clauses to $\sim$56K on a question, while our method, described next, brought it down to only 951 clauses. Further, Lifted Blocked Gibbs and Probabilistic Theorem Proving, as implemented in Alchemy-2, were slower than basic Alchemy-1.

Different from heuristic approximations 
%(e.g., lazy grounding in Tuffy~\cite{tuffy2011} and the 
(e.g., Modified Closed-World Assumption of \namecite{beltagy2014efficient}), we propose reducing
the grounding size without altering
the semantics of the MLN program. We utilize the combinatorial
structure imposed by the set $H$ of hard constraints present in the MLN,
and use it to simplify the grounding of \emph{both} hard and soft constraints.
%The approach thus embraces hard clauses rather than relaxing them, as is often done in probabilistic inference techniques, especially when avoiding infinite energy barriers in MCMC based methods.
Lazy inference mentioned above may be viewed as the very first step
of our approach.

Assuming an arbitrary ordering of the constraints in $H$, let $F_i$
denote the first $i$ constraints. Starting with $i=1$, we
generate the propositional grounding $G_i$ of $F_i$, use a
propositional satisfiability (SAT) solver to identify the set $B_i$ of
\emph{backbone variables} of $G_i$ (i.e., variables that take a fixed
value in all solutions to $G_i$), freeze values of the corresponding
atoms in $B_i$, increment $i$, and repeat until $G_{|H|}$ has been
processed. Although the end result can be described simply as
freezing atoms corresponding to the backbone
variables in the grounding of $H$, the incremental process helps us
keep the intermediate grounding size under control as a propositional
variable is no longer generated for an atom once its value is
frozen. Once the freezing process is complete, the full grounding of
$H$ is further simplified by removing frozen variables. Finally, the
soft constraints $S$ are grounded much more efficiently by taking
frozen atoms into account.

This approach can be seen as an extension of a proposal by
\namecite{papai2011constraint}.
%Rather than identifying all backbone
%variables through $2 N_i$ calls to a complete SAT solver (where $N_i$
%is the number of variables of $F_i$),
They used a polynomial-time Generalized Arc Consistency
%(specifically, linear-time unit propagation)
algorithm on $H$ to compute a subset of
$B_{|H|}$ that is efficiently identifiable,
%. Further, while their
%implementation
implemented as a sequence of \emph{join} and \emph{project}
database operations
%to implement Unit Propagation on the fly. while
%grounding, we exploit a state-of-the-art SAT solver on the base ground
%network for each $i$.
%for performing Unit Propagation as well as the
%coNP-hard task of backbone detection.

% Lazy Inference, which identifies the set of active ground constraints
% and atoms by freezing the values of only the hard evidence atoms, can
% be viewed as the very first step of our approach with unit propagation,
% which in turn is subsumed by the backbone method. 
%  %As a result, our approach results in the same or fewer number of
%  %groundings than lazy inference.
% For instance, lazy inference in Alchemy-1 reduced $\sim$70K ground clauses to $\sim$56K on a question, while our technique brought it down to only 951 ground clauses. Further, Lifted Blocked Gibbs and Probabilistic Theorem Proving, as implemented in Alchemy-2, were slower than basic Alchemy-1 inference.
%%%%%%%%%%%%%%%%%%%%%
%% CONSTANTS BASED MLN / ENTITY RESOLUTION
%%%%%%%%%%%%%%%%%%%%%
%\vspace{-0.5em}
\subsection{(B) Entity Resolution Based MLN}

Representing generalities as quantified rules defined over classes of entities or events appears to be a natural formulation, but is also quite inefficient leading to large grounded networks. We instead consider an alternative formulation that treats generalities as relations expressed over \emph{prototypical} entities and events. This formulation leverages the fact that elementary level science questions can often be answered using relatively simple logical reasoning over exemplar objects and homogeneous classes of objects, if given perfect information as input. The only uncertainty present in our system is what's introduced by lexical variations and extraction errors, which we handle with probabilistic equality.

{\bf KB Rules and Question}: We create rules defined over prototypical entity/event {\em constants}, rather than first-order variables.  These constants are tied to their respective string representations, with the understanding that two entities behave similarly if they have lexically similar strings.  E.g.,
\begin{align*}
  & \isa(G,\text{grow}), \isa(A,\text{some\_animals}), \isa(F,\text{thicker\_fur}),\\
  & \isa(W,\text{the\_winter}), \agent(G,A), \object(G,F), \mathit{in}(G,W)\\
  & \Rightarrow \isa(S,\text{stays}), \isa(R,\text{warm}),\\
  & \ \ \ \ \ \ \ \enables(G,S), \agent(S,A), \object(S,R)
\end{align*}
What was a first-order rule in FO-MLN is now already fully grounded! It has no variables.  Entities/events mentioned in the question are also similarly represented by constants.

{\bf Equivalence or Resolution Rules}: Using a simple probabilistic variant of existing Entity/Event Resolution frameworks \cite{singla2006er,kok2008er}, we ensure that (a) two entities/events are considered similar when they are tied to lexically similar strings and (b) similar entities/events participate in similar relations w.r.t.\ other entities/events. This defines soft \emph{clusters} or equivalence classes of entities/events.  To this end, we use a probabilistic $\sameas$ predicate which is reflexive, symmetric, and transitive, and interacts with the rest of the MLN as follows:
\begin{align*}
w(s,s'):\ \entails(s,s') & \\
  \isa(x,s), \entails(s,s') & \to \isa(x,s').\\
  \isa(x,s), \isa(y,s) & \to \sameas(x,y).\\
w:\ \isa(x,s), !\isa(y,s) & \to \ !\sameas(x,y)\\
  r(x,y), \sameas(y,z) & \to r(x,z).
\end{align*}
$r$ in the last rule refers to any of the MLN predicates other than $\entails$ and $\isa$. The $\sameas$ predicate, as before, is implemented in a typed fashion, separately for entities and events. We will refer to this formulation as ER-MLN.

{\bf Partial Match Rules} Due to lexical variability, often not all conjuncts in a rule's $\antecedent$ are present in the question's $\setup$. To handle incomplete matches, for each KB derived MLN rule of the form $(\wedge_{i=1}^k L_i) \to R$, we also add $k$ soft rules of the form $L_i \to R$. This adds flexibility, by helping ``fire'' the rule in a soft manner.

{\bf Comparison with FO-MLN}: Long KB rules and question representation now no longer have quantified variables, only the binary or ternary rules above do. These mention at most $3$ variables each and thus have relatively manageable groundings. On the other hand, as discussed in the next section, ER-MLN can fail on questions that have distinct entities with similar string representations.
Further, it is brittle to syntactic differences such as $\agent(\text{Fall},\text{Things})$ generated by ``things fall due to gravity'' and $\object(\text{Dropped},\text{Ball})$ for ``a student dropped a ball''. Although ``drop'' entails ``fall'' and ``ball'' entails ``object'', ER-MLN cannot reliably bridge the structural difference involving $\object$ and $\agent$, as these two relationships typically aren't equivalent. %On many questions, such differences result in missed opportunities in applying relevant KB rules.
Despite these limitations, ER-MLN provides a substantial scalability advantage over FO-MLN on a vast majority of the questions that remain within its scope.

\subsection{(C) PRobabilistic ALignment and INferencE}
%While ER-MLN can handle partially missing edges by breaking the KB rule, it still relies primarily on the predicates (also referred to as links or edges) for inference. %While our inputs don't capture this structural variation in text,\footnote{Resources such as PPDB \cite{ganitkevitch2013ppdb} may be used to obtain such information, but they are likely to introduce significant noise as well.} we already capture the lexical variation using the entailment score. 
ER-MLN handles some of the word-level lexical variation via {\em resolution} and soft {\em partial match} rules that break long antecedents. However, it is still rigid in two respects:

\begin{enumerate}

\item Inference primarily relies on the predicates (also referred to as links or edges) for inference. As a result, even if the words in the $\antecedent$ and $\setup$ have high entailment scores, the rule will still not ``fire'' if the edges do not match. To enable effective rule application under such circumstances, we require (at a minimum) some match on the string constants and allow edge matches (if any) to increase the confidence in rule application.

\item As clustering forces entities bound to lexically equivalent strings to ``behave'' identically, it fails on questions that involve two different entities that are bound to equivalent string representations. %Consider the question: ``A student puts two identical plants in the same type and amount of soil. She puts one of these plants near a sunny window and the other in a dark room. This experiment tests how the plants respond to (A) light (B) air (C) water (D) soil.''  The entities corresponding to the two plants will be bound to equivalent string representations and hence will be treated as the same entity.
To avoid this issue, we do not force the entailment-based clusters of constants to behave similarly. Instead, as we discuss below, we use the clusters to guide inference in a softer manner.

\end{enumerate}

%Text-derived rules present many types of variability. ER-MLN handles some of the word-level lexical variation via {\em resolution} and soft {\em partial match} rules that break long antecedents in the KB rule. However, it still primarily relies on the predicates (also referred to as links or edges) for inference. As a result, even if the words in the $\antecedent$ and $\setup$ have high entailment scores, the rule will still not fire if the edges do not match. To enable effective rule application under such circumstances, we require (at a minimum) some match on the string constants and allow edge matches (if any) to increase the confidence in rule application.

To introduce such flexibility, we convert the problem of uncertainty over links between string constants to the problem of uncertainty over the existence of these constants.  To this end, we introduce a unary predicate over string constants to capture what is known to be true (i.e., the $\setup$) or can be proven to be true (via inference using the KB rules) in the world. We then define an MLN to directly control how new facts are inferred given the KB rules. 
%tushar: Moved it up into the ER problem and solutions list. --> Moreover, we do not force the entailment-based constant clusters to behave similarly. Instead, as we show below, we use the clusters to guide inference. 
%Defining an MLN that directly controls how we infer the \emph{existence} of new constants gives us 
The flexibility to control inference helps address two additional QA challenges:
%\begin{enumerate}

{\bf Acyclic inference}: While knowledge is extracted from text as a set of directed rules each with an $\antecedent$ and a $\consequent$, there is no guarantee that the rules taken together are acyclic. E.g., a rule stating ``Living things $\to$ depend on the sun'' and ``Sun $\to$ source of energy for living things'' may exist side-by-side. Successful inference for QA must avoid feedback loops.

{\bf False unless proven}: While MLNs assume atoms not mentioned in any rule to be true with probability 0.5, elementary level science reasoning is better reflected in a system that assumes all atoms to be false unless stated in the question or proven through the application of a rule. This is similar to the semantics of Problog \cite{problog} and PRISM \cite{prism}. %These semantics are also captured by the modified closed world assumption of \namecite{beltagy2014efficient}.

%\end{enumerate}

% \niranjan{Say some more about the existence uncertainty to understand how the grounding size reduces. It maybe better to move this to the end of this section.}
%\tushar{Makes sense and moved here. I added additional details above to make the \emph{flexbility} argument}. 

% \begin{figure*}[tb]
% \centering
% \includegraphics[trim=0px 20px 10px 190px, clip, scale=0.5]{figures/query-rule.pdf}
% \caption{KB rule and question as a graph. $\setup$ is blue, $\query$ is green, $\antecedent$ is orange, and $\consequent$ is purple.}
% \label{arilog_graph}
% \end{figure*}

\begin{SCfigure*}
\centering
\includegraphics[trim=0px 10px 0px 160px, clip, scale=0.5]{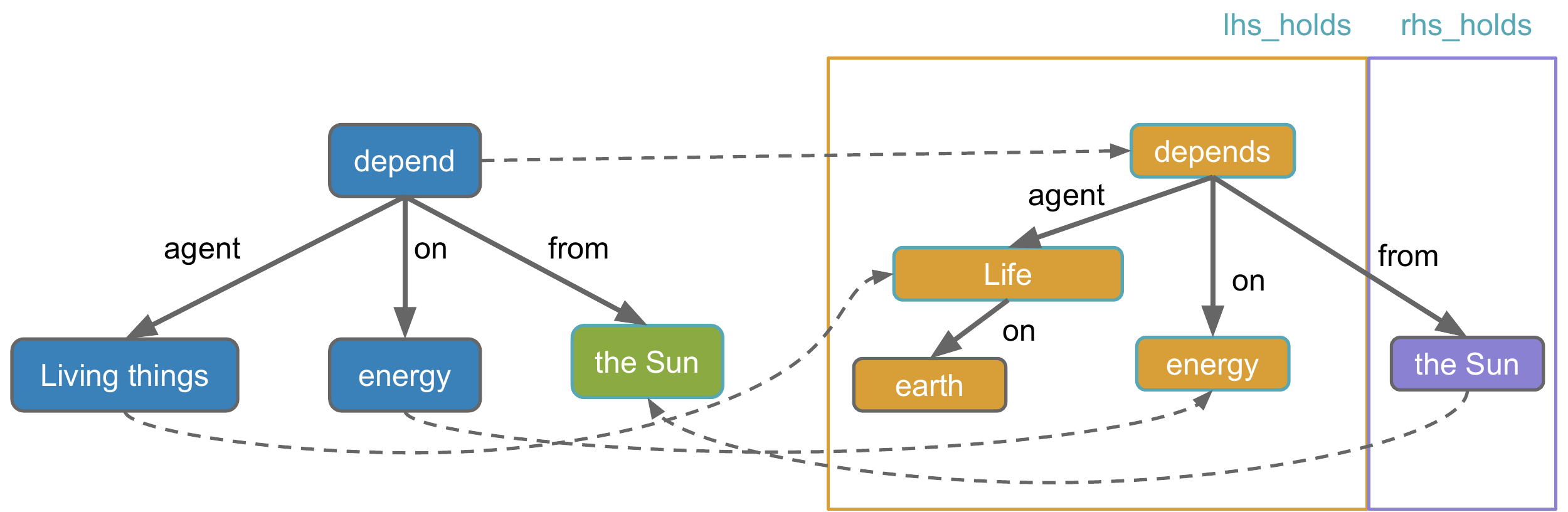}
\caption{KB rule and question as a graph. $\setup$ is blue, $\query$ is green, $\antecedent$ is orange, and $\consequent$ is purple. Alignments are shown with dotted lines. $\lhsholds$ combines the individual probabilities of $\antecedent$ nodes.}
\label{alignments}
\end{SCfigure*}

\textbf{MLN specification}
We introduce a unary predicate called $\holds$ over string constants to capture the probability of a string constant being true given the $\setup$ is true ($\forall x \in \setup, \holds(x)=\mathit{true}$) and the KB rules hold. Instead of using edges for inference, we use them as factors influencing alignment: similar constants have similar local neighborhoods. This reduces the number of unobserved groundings from $O(n^2)$ edges in the ER-MLN to $O(n)$ existence predicates, where $n$ is the number of string constants. 
%While our previous approaches can handle the lexical variation and incomplete matches, they primarily rely on matching the edges between the KB rule and the question for inference. 
For the example rule (\ref{rule:fox}), Praline can be viewed as using the following rule for inference:
% ER-MLN effectively (either as a single rule or multiple sub-rules) uses the implication:
% \begin{align*}
% & \agent(\mathit{Grow}, \mathit{Animals}), \object(\mathit{Grow}, \mathit{Fur}), \\
% & \mathit{in}(\mathit{Grow}, \mathit{Winter}) \Rightarrow \enables(\mathit{Grow}, \mathit{Stays}), \\
% & \quad \agent(\mathit{Stays}, \mathit{Animals}), \object(\mathit{Stays}, \mathit{Warm})
% \end{align*}
% We instead modify this to:
\begin{align*}
& \holds(\mathit{Grow}), \holds(\mathit{Animals}), \holds(\mathit{Fur}), \\
& \holds(\mathit{Winter}) \Rightarrow \holds(\mathit{Stays}), \holds(\mathit{Warm}) 
\end{align*}
%Instead of transforming the dependency structure into first-order logic, we directly use the graph structure for inference. Each rule can be viewed as introducing additional graph structure ($consequent$) based on the match with the $\antecedent$ graph. The question-answering task can now be viewed as sequentially applying rules to the $\setup$ to best match the $\query$. 
%To perform the matching of the $\antecedent$ and the $setup$, we use soft (probabilistic) graph matching defined using MLN clauses, similar to the \emph{ConstMLN} approach. To perform inference using multiple rules over these alignments, we define a novel MLN that efficiently handles the issues described before and two additional issues introduced by chaining multiple rules which are not handled in our other formulations. %it introduces two additional To ensure acylicity of inference, we define hard rules that prevent any sub-graph to be true without being derived from the $\setup$. We next describe each set of rules. 
%\tushar{TO ADD: Example of Arilog - Graph - Facts}
%The question and rule representation can be viewed as a graph shown in Fig~\ref{arilog_graph}
If we view KB rules and the question as a labeled graph $G$ shown in Figure \ref{alignments}, alignment between string constants corresponds to alignment between the nodes in $G$. The nodes and edges of $G$ are the input to the MLN, and the $\holds$ predicate on each node captures the probability of it being true given the $\setup$. We now use MLNs (as described below) to define the inference procedure for any such input graph $G$. 

%% Ashish: \paragraph is nicer, but changing to \textbf to align with everything else earlier.
%\paragraph{Input Predicates: }
\textbf{Input Predicates}:
We represent the graph structure of $G$ using predicates $\node(nodeid)$ and $\edge(\mathit{nodeid}, \mathit{nodeid}, \mathit{label})$. We use $\setup(\mathit{nodeid})$ and $\query(\mathit{nodeid})$ to represent the question's $\setup$ and $\query$, resp. Similarly, we use $\inlhs(\mathit{nodeid})$ and $\inrhs(\mathit{nodeid})$ to represent rules' $\antecedent$ and $\consequent$, resp. 

%\paragraph{Graph Alignment Rules: }
\textbf{Graph Alignment Rules}:
Similar to the previous approaches, we use entailment scores between words and short phrases to compute the alignment. In addition, we also expect \emph{aligned} nodes to have similar edge structures:
\begin{align*}
\aligns(x, y), \edge(x, u, r), \edge(y, v, s) \Rightarrow \aligns(u, v) \\
\aligns(u, v), \edge(x, u, r), \edge(y, v, s) \Rightarrow \aligns(x, y)
\end{align*}
That is, if node $x$ aligns with $y$ then their children/ancestors should also align. We create copies of these rules for edges with the same label, $r = s$, with a higher weight and for edges with different labels, $r \neq s$, with a lower weight.

%\paragraph{Inference Rules: } 
\textbf{Inference Rules}: 
%To denote nodes that are true or proven to be true, we introduce a $holds(nodeid)$ predicate which is set to true for the setup nodes and unobserved for all the other nodes. We use the inference rules to prove additional nodes and finally compute the query probability using the $\holds$ probability of the query nodes.
%
We use MLNs to define the inference procedure to prove the $\query$ using the alignments from $\aligns$. We assume that any node $y$ that $\aligns$ with a node $x$ that $\holds$, also $\holds$:
\begin{align}
\label{eq:1}
\holds(x), \aligns(x, y) \Rightarrow \holds(y) 
\end{align}

For example, if the setup mentions ``fox", all nodes that entail ``fox'' also hold. As we also use the edge structure during alignment, we would have a lower probability of ``fox" in ``fox finds food" to align with ``animal" in ``animal grows fur" as compared to ``animal" in ``animal finds food".

We use KB rules to further infer new facts that should hold based on the rule structure. We compute $\lhsholds$, the probability of the rule's $\antecedent$ holding, and use it to infer $\rhsholds$, the probability of the $\consequent$. Similar to ER-MLN, we break the rule into multiple small
%rules to combine the $\holds$ probability of each node.
rules.\footnote{An intuitive alternative for the 2nd clause doesn't capture the intending meaning, $-w: !\holds(x), \inlhs(x, r) \Rightarrow \lhsholds(r)$}
%Since our KB rules can be very long, instead of using a single MLN rule $\wedge_i \holds(x_{ij}) \Rightarrow \lhsholds(r_j)$ that can potentially slow down inference, we use two short rules to combine the $\holds$ probability of each node.\footnote{An intuitive alternative for the 2nd clause doesn't capture the intending meaning, $-w: !\holds(x), \inlhs(x, r) \Rightarrow \lhsholds(r)$}
%
\begin{align*}
w: & \holds(x), \inlhs(x, r) \Rightarrow \lhsholds(r) \\
w: & !\holds(x), \inlhs(x, r) \Rightarrow !\lhsholds(r) \\
& \lhsholds(r) \Rightarrow \rhsholds(r). \\
& \rhsholds(r), \inrhs(r, x) \Rightarrow \holds(x).
\end{align*}
\textbf{Acyclic inference}:
We use two predicates, $\proves(\mathit{nodeid}, \mathit{nodeid})$ and $\ruleProves(\mathit{rule}, \mathit{rule})$ to capture the inference chain between nodes and rules, resp. We can now ensure acyclicity in inference by introducing transitive clauses over these predicates and disallowing reflexivity, i.e., $!\proves(x, x)$.
%, e.g.,
%\begin{align*}
%\proves(x, y), \proves(y, z) \Rightarrow \proves(x, z).\\
%!\proves(x, x).
%\end{align*}
 We update rule (\ref{eq:1}) to:
\begin{align*}
w: & \proves(x, y), \holds(x) \Rightarrow \holds(y) \\
w: & \aligns(x, y) \Rightarrow \proves(x, y)
\end{align*}

We capture the direction of inference between rules by checking consequent and antecedent alignments:
\begin{align*}
& \proves(x, y), inrhs(x, r1), inlhs(y, r2) \Rightarrow  \ruleProves(r1, r2).
\end{align*}
\textbf{False unless proven}:
To ensure that nodes hold only if they can be proven from $\setup$, we add bidirectional implications to our rules. An alternative is to introduce a strong negative prior on $\holds$ and have a higher positive weight on all other clauses that conclude $\holds$. However, the performance of our MLNs was very sensitive to the choice of the weight. We instead model this constraint explicitly. Figure \ref{alignments} shows a sample inference chain using dotted lines.  %Due to lack of space, we refer the reader to \tushar{INSERT URL} for the complete MLN.

%We achieve the former by checking for cycles in the proof paths. We achieve the latter by checking for both the if and only if conditions in our MLN clauses. 

% using the rule and question from Figure \ref{arilog_graph}. 
 Praline defines a meta-inference procedure that is easily modifiable to enforce desired QA inference behavior, e.g. $w: \aligns(x, y), \setup(x) \Rightarrow !\query(y)$ would prevent a term from the $\setup$ to align with the $\query$. Further, by representing the input KB and question as evidence, we can define a single static first-order MLN for all the questions instead of a compiled MLN for every question. This opens up the possibility of learning weights of this static MLN, which would be challenging for the previous two approaches.\footnote{In this work, we have set the weights manually.} 

\begin{table*}[tb]
\centering
\caption{QA performance of various MLN formulations. The number of MLN rules, number of ground clauses, and runtime per multiple-choice question are averages over the corresponding dataset. \#Answered column indicates questions where at least one answer option didn't time out (left) and where no answer option timed out (right). \#Atoms and \#GroundClauses for FO-MLN are averaged over the 398 MLNs where grounding finished; 34 remaining MLNs timed out after processing 1.4M clauses.}
\label{table:mln-comparison}
\small
\setlength{\tabcolsep}{8pt}
\setlength{\doublerulesep}{\arrayrulewidth}
\begin{tabular}{c c | c c | c c c | c } \toprule
Question & MLN & \#Answered & Exam & \#MLN & \#Atoms & \#Ground & Runtime \\
Set & Formulation & (some / all) & Score & Rules & & Clauses & (all) \\
\hline \toprule
\multirow{3}{*}{Dev-108}
 & FO-MLN & 106 / 82 & 33.6\% & 35 & 384$^\ast$ & 524$^\ast$ & 280 s \\
 & ER-MLN & 107 / 107 & 34.5\% & 41 & 284 & 2,308 & 188 s \\
 & PRALINE & 108 & {\bf 48.8\%} & 51 & 182 & 219 & {\bf 17 s} \\
\toprule
\multirow{3}{*}{Unseen-68}
 & FO-MLN & 66 & 33.8\% & - & - & - & 288 s \\
 & ER-MLN & 68 & 31.3\% & - & - & - & 226 s \\
 & PRALINE & 68 & {\bf 46.3\%} & - & - & - & {\bf 17 s} \\
\hline
\end{tabular}
\end{table*}
%%%%%%%%%%%%%%%%%%%%%
%% EXPERIMENTS
%%%%%%%%%%%%%%%%%%%%%
\vspace{-0.5em}
\section{Empirical Evaluation}
%To evaluate our three MLN formulations of the QA task,
We used Tuffy 0.4\footnote{{http://i.stanford.edu/hazy/tuffy}}
\cite{tuffy2011} as the base MLN solver\footnote{We also tried Alchemy 1.0, 
%({\scriptsize\url{http://alchemy.cs.washington.edu}}),
which gave similar results.} and extended it to incorporate
the hard-constraint based grounding reduction technique discussed
earlier, implemented using the SAT solver Glucose
3.0\footnote{{http://www.labri.fr/perso/lsimon/glucose}}
\cite{glucose2009} exploiting its ``solving under assumptions'' capability for efficiency.
%For each true-false question, we gave Tuffy 
We used a 10 minute timelimit, including a max of 6 minutes for grounding.
%up to 6 minutes to generated the grounding and up to 10 minutes total. 
Marginal inference was performed using MC-SAT~\cite{mcsat2006}, with default parameters and 5000 flips per sample to generate 500 samples for marginal estimation.
% Eric: so actually we also set the inverse of the SA temp to 2, instead of the default 10...not sure this parameter was even used in the default Tuffy, but the config option is hard-coded into solvers/mln-common/src/main/scala/org/allenai/ari/mlncommon/tuffy/TuffyExecutor.scala
% except for turning XYZ (???)  off, 
%The SAT solver used for unit
% propagation and backbone detection was Glucose
% 3.0\footnote{{http://www.labri.fr/perso/lsimon/glucose}}
% \cite{glucose2009}. We used Glucose's ``solving under \emph{assumptions}''
% feature to improve the efficiency of backbone detection by sharing
% learned clauses across $2 N$ calls to the main search routine for
% a network with $N$ ground atoms.

We used a 2-core 2.5 GHz Amazon EC2 linux machine with
%2.5 GHz Intel Xeon E5-2670 v2  processor and
16 GB RAM.
%running Ubuntu Linux. %Starting with the \emph{Aristo Science Exams}
%dataset\footnote{\url{http://allenai.org/TemplateGeneric.aspx?contentId=17}}
We selected 108 elementary-level science questions (non-diagram, multiple-choice) from 4th grade New
York Regents exam as our benchmark (Dev-108) and used another 68 questions as a blind test set (Unseen-68).\footnote{Question sets, MLNs, and our modified Tuffy solver are available at {\small \url{http://allenai.org/software.html}}}
% We used the study guide and a small set of sentences 
% automatically retrieved from the web to improve coverage of the knowledge 
% in relation to the questions. NLP techniques were used to extract
% knowledge from these sentences, which was then formulated as MLN rules
% as discussed earlier. 

The KB, representing roughly 47,000 sentences, was generated in advance by processing the New York Regents 4th grade science exam syllabus, the corresponding Barron's study guide, and documents obtained by querying the Internet for relevant terms.  Given a question, we use a simple word-overlap based matching algorithm, referred to as the \emph{rule selector}, to retrieve the top $30$ matching sentences to be considered for the question.
Textual entailment scores between words and
short phrases were computed using WordNet~\cite{miller1995wordnet}, and
converted to ``desired" probabilities for the corresponding soft $\entails$
evidence. The accuracy reported for each approach is computed as the
number of multiple-choice questions it answers correctly, with a
partial credit of $1/k$ in case of a $k$-way tie between the highest
scoring options if they include the correct answer.

%%%%%%%%%%%%
\subsection{MLN Formulation Comparison}

Table~\ref{table:mln-comparison} compares the effectiveness of our three MLN formulations, named FO-MLN, ER-MLN, and Praline. For each question and each approach, an MLN program was generated for each of the answer options using the most promising KB rule for that answer option.

In the case of FO-MLN, Tuffy exceeded the 6 minute time limit when generating groundings for 34 of the $108 \times 4$ MLNs for the Dev-108 question set, quitting after working with $1.4 \times 10^6$ clauses on average, despite starting with only around 35 first-order MLN rules. In the remaining MLNs, where our clause reduction technique successfully finished, there is a dramatic reduction in the ground network size: only 524 clauses and 384 atoms on average.

Tuffy finished inference for all 4 answer options for 82 of the 108 questions; for other questions, it chose the most promising answer option among the ones it finished processing. Overall, this resulted in a score of 33.6\% with an average of 280 seconds per multiple-choice question on Dev-108, and similar performance on Unseen-68.

ER-MLN, as expected, did not result in any timeouts during grounding. The number of ground clauses here, 2,308 on average, is dominated not by KB rules but by the binary and ternary entity resolution clauses involving the $\sameas$ predicate. ER-MLN was roughly 33\% faster than FO-MLN, but overall achieved similar exam scores as FO-MLN.

Praline resulted in a 10x speedup over ER-MLN, explained in part by much smaller ground networks with only 219 clauses on average. Further, it boosted exam performance by roughly 15\%, pushing it up to 48.8\% on Dev-108 and 46.3\% on Unseen-68. This demonstrates the value that the added flexibility and control of Praline bring.
%to MLN formulations of QA.

%%%%%%%%%%%%
\subsection{Praline: Improvements and Ablation}

\begin{table}[tb]
\centering
\caption{QA performance of Praline MLN variations.}
\label{table:praline-comparison}
\small
\setlength{\tabcolsep}{5pt}
\setlength{\doublerulesep}{\arrayrulewidth}
\begin{tabular}{l | c c | c c} \toprule
  & \multicolumn{2}{c|}{One rule} & \multicolumn{2}{c}{Chain$=$2} \\
\multicolumn{1}{c|}{MLN} & Dev-108 & Unseen & Dev-108 & Unseen \\
\hline \toprule
Praline & 48.8\% & 46.3\%& {\bf 50.3\%} & {\bf 52.7\%} \\
\hline
No Acyclic & 44.7\% & 36.0\% & 43.6\% & 30.9\% \\
No FUP & 35.0\% & 30.9\% &  42.1\% & 29.4\% \\
No FUP, Acyclic & 37.3\% & 34.2\% & 36.6\% & 24.3\% \\
\hline
\end{tabular}
\end{table}

We next compare the performance of Praline when using multiple KB rules as a chain or multiple inference paths. Simply using the top two rules for inference turns out to be ineffective as the top two rules provided by the rule selector are often very similar. Instead, we use \emph{incremental inference} where we add one rule, perform inference to determine which additional facts now hold and which $\setup$ facts haven't yet been used, and then use this information to select the next best rule. This, as the Chain$=$2 entries in the first row of Table~\ref{table:praline-comparison} show, improves Praline's accuracy on both datasets. The improvement comes at the cost of a modest increase in runtime from 17 seconds per question to 38.

Finally, we evaluate the impact Praline's rules for handling acyclicity (Acyclic) and the false-unless-proven (FUP) constraint. Table \ref{table:praline-comparison} shows a drop in Praline's accuracy when either of these constraints is removed,
%when we remove the rules enforcing the false unless proven constraint or acyclicity rules from the original formulation. 
which highlights their importance in our QA task. Specifically, when we use only one KB rule, dropping FUP clauses has a larger influence on the score as compared to dropping the Acyclic constraint. Removing Acyclic constraint still causes a small drop even with a single rule due to the possibility of cyclic inference within a rule. When chaining multiple rules, however, cyclic inference becomes more likely and we see a correspondingly larger reduction in score when dropping Acyclic constraints.

%%%%%%%%%%%%%%%%%%%%%
%% DISCUSSION / CONCLUSION
%%%%%%%%%%%%%%%%%%%%%
\section{Discussion}

Our investigation of the potential of MLNs for QA resulted in multiple formulations, the third of which is a flexible model that outperformed other, more natural approaches. We hope our question sets and MLNs will guide further research on improved modeling of the QA task and design of more efficient inference mechanisms for such models. 

While SRL methods seem a perfect fit for textual reasoning tasks such as RTE and QA, their performance on these tasks is still not up to par with simple textual feature-based approaches \cite{beltagy2014efficient}. On our datasets too,
%the average graph-matching score between the $\setup$-$\antecedent$ and $\consequent$-$\query$ as a confidence measure achieves an accuracy of 57.8\% on the Dev set.
simple word-overlap based approaches perform quite well, scoring around 55\%.
%\tushar{I wrote it but not sure if I should make this claim $\rightarrow$?}
We conjecture that the increased flexibility of complex relational models comes at the cost of increased susceptibility to noise in the input. Automatically learning weights of these models may allow leveraging this flexibility in order to handle noise better. Weight learning in these models, however, is challenging as we only observe the correct answer for a question and not intermediate feedback such as ideal alignments and desirable inference chains.

Modeling the QA task with MLNs, an undirected model, gives the flexibility to define a joint model that allows alignment to influence inference and vice versa. At the same time, inference chains themselves need to be acyclic, suggesting that models such as Problog and SLP would be a better fit for this sub-task. Exploring hybrid formulation and designing more efficient and accurate MLNs or other SRL models for the QA task remains an exciting avenue of future research. 
%%%%%%%%%%%%%%%%%%%%%
%% ACK
%%%%%%%%%%%%%%%%%%%%%
\subsection{Acknowledgments}
\begin{small}
The authors would like thank Pedro Domingos and Dan Weld for invaluable discussions
%regarding efficient MLN formulations,
and the Aristo team at AI2, especially Jesse Kinkead, for help with prototype development and evaluation.
\end{small}

%%%%%%%%%%%%%%%%%%%%%
%% BIBLIOGRAPHY
%%%%%%%%%%%%%%%%%%%%%

\bibliographystyle{aaai}
%\begin{small}
\bibliography{mln-qa}
%\end{small}

\end{document}